\documentclass[letterpaper, 10 pt, conference]{ieeeconf}  % Comment this line out if you need a4paper

\IEEEoverridecommandlockouts                              % This command is only needed if 
\usepackage{graphics} % for pdf, bitmapped graphics files
\usepackage{epsfig} % for postscript graphics files
\usepackage{mathptmx} % assumes new font selection scheme installed
\usepackage{times} % assumes new font selection scheme installed
\usepackage{amsmath} % assumes amsmath package installed
\usepackage{amssymb}  % assumes amsmath package installed
\usepackage{xcolor}
\usepackage{multirow}
\usepackage[colorlinks=true]{hyperref}

\title{\LARGE \bf Perceive, Attend, and Drive: \\Learning Spatial Attention for Safe Self-Driving}

\author{
Bob Wei$^{*,1,2}$, Mengye Ren$^{*,1,3}$, Wenyuan Zeng$^{1, 3}$, Ming Liang$^{1}$, Bin Yang$^{1,3}$, Raquel Urtasun$^{1, 2}$%
\thanks{$^{*}$Equal contribution $^{1}$Uber ATG $^{2}$University of Waterloo $^{3}$University of Toronto. Correspondence: {\tt\small q25wei@uwaterloo.ca, \{mren, urtasun\}@cs.toronto.edu}}}

\DeclareMathOperator*{\argmin}{arg\,min}

\newcommand{\attn}{A}
\newcommand{\temp}{K}
\newcommand{\sattn}{\tilde{A}}

\newcommand{\cC}{C}
\newcommand{\cL}{L}
\newcommand{\cLcla}{L_{\text{cls}}}
\newcommand{\cLreg}{L_{\text{reg}}}
\newcommand{\cLclaij}{L_{\text{cls},i,j}}
\newcommand{\cLregij}{L_{\text{reg},i,j}}
\newcommand{\cLpln}{L_{\text{plan}}}
\newcommand{\cLatn}{L_{\attn}}

\newcommand{\lcla}{\lambda_{\text{cls}}}
\newcommand{\lreg}{\lambda_{\text{reg}}}
\newcommand{\lpln}{\lambda_{\text{plan}}}
\newcommand{\lalp}{\lambda_{A}}

\newcommand{\Inp}{X}

\DeclareMathOperator*{\sigmoid}{\mathrm{sigmoid}}

\newcommand{\ourdata}{Drive4D}

\begin{document}

\maketitle
\thispagestyle{empty}
\pagestyle{empty}
%%%%%%%%%%%%%%%%%%%%%%%%%%%%%%%%%%%%%%%%%%%%%%%%%%%%%%%%%%%%%%%%%%%%%%%%%%%%%%%%

% !TEX root = ../main.tex
\begin{abstract}
% Inspired by the usage of visual attention in the human brain, 
In this paper, we propose an end-to-end self-driving network featuring a sparse attention module that learns to automatically attend to important regions of the input. The attention module specifically targets motion planning, whereas prior literature only applied attention in perception tasks. Learning an attention mask directly targeted for motion planning significantly improves the planner safety by performing more focused computation. Furthermore,  visualizing the attention  improves interpretability of end-to-end self-driving.
\end{abstract}

% !TEX root = ../main.tex

\section{Introduction}
Self-driving is one of today's most impactful technological challenges, one that promises to bring
safe and affordable  transportation everywhere. Tremendous improvements have been made
in self-driving perception systems, thanks to the success of deep learning. This has enabled
accurate detection and localization of obstacles, providing a holistic understanding of the
surrounding world, which is then sent to the motion planner to decide subsequent driving actions.

\begin{figure}[t]
  \centering
  \includegraphics[width=\linewidth]{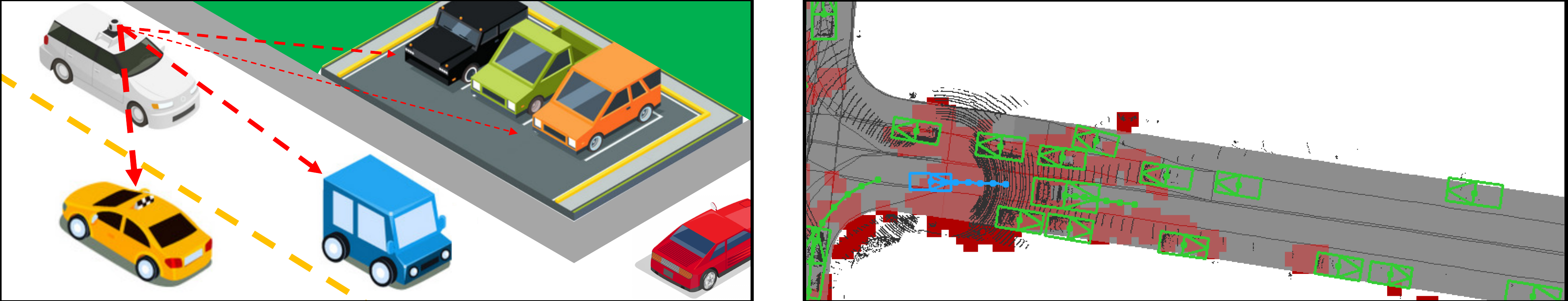}
  \caption{\small \textit{Left:} a toy example. \textit{Right:} our learned spatial attention in red with ego vehicle in blue and others in green. Not all actors impact our safe driving and so we should prioritize accordingly.}
  \label{fig:toy}
  \vspace{-0.15in}
\end{figure}

Despite the eminent success of these perception systems, their detection
objective is \textit{mis-aligned} with the self-driving vehicle's overall goal---to drive
safely to the destination. We typically train the perception systems to detect all objects in
the sensor range, assigning each object an equal weight even if some objects are not important
as they will never interact with the self-driving vehicle. For example, they could be far away or
parked on the other side of the road as in Figure~\ref{fig:toy}.
As a result, a vast amount of computation and model capacity is wasted in recognizing
very difficult instances that matter only for common metrics such as average precision (AP), but not
so much for driving.
This is in
striking contrast with how humans drive: we focus our visual attention in areas that directly impact
safe planning. 
Inspired by the use of visual attention in our brain, we aim to introduce attention to self-driving systems to efficiently and selectively process complex scenes.

Numerous studies in the past have explored adding sparse attention in deep neural networks to
improve computation efficiency in  classification~\cite{resattn,adaptivecomp} and object
detection~\cite{sbnet,progressivesparse,recattend}. 
In order to perform well on the metrics employed in common benchmarks, the
attention mask in~\cite{sbnet} still needs to cover all actors in the scene, slowing the network when the scene has many vehicles.

In this paper, our aim is to address these inconsistencies such that the amount of computation is optimized towards the end goal of motion planning.
Specifically
our contributions are as follows:
\begin{itemize}
\item We learn an attention mask directly towards the motion planning objective for safe self-driving.
\item We use the attention mask to reweight object detection and motion forecasting losses in our
joint end-to-end training, focusing the model capacity on objects that matter most.
Different from manually prioritizing instances~\cite{prioritize}, here the weighting is entirely
data-driven.
\item Our attention-based model significantly reduced the
collision rate and improved planning performance while at a
much lower computation cost.
\item Attention mask visualization improves interpretability of end-to-end deep learning
models in self-driving.
\end{itemize}

% !TEX root = ../main.tex
\section{Related Work}
\textbf{Attention mechanism in deep learning:}
Human and other primate visual perception systems feature visual attention to reduce the complexity
of the scene and speed-up inference~\cite{neurobiology,saliencyvisattend}. Earlier studies in visual
saliency aimed to predict human gaze with no particular task in mind~\cite{predicthuman}. Attention
mechanisms nowadays are built in as  part of  end-to-end models to optimize towards specific tasks.
The attention modules are typically implemented as multiplicative gates to select features. This
schema has shown to improve performance and interpretability on downstream tasks such as object
recognition~\cite{visattend,attendrbm,resattn}, instance segmentation~\cite{recattend}, image
captioning~\cite{showattendtell}, question answering~\cite{coattend,san}, as well as other natural language processing applications~\cite{machinetrans,transformer,bert}. The visualization of
the end-to-end learned attention suggests that deep attention-based models have an intelligent
understanding of the inputs by focusing on the most informative parts of the input.

\textbf{Sparse activation in neural networks:}
Sparse coding models~\cite{sparsecoding} use an
overcomplete dictionary to achieve sparse activation in the feature space. In modern convolutional neural networks (CNNs), sparsity
is typically brought by the widespread use of ReLU activation functions, but these
are rather unstructured, and speed-up has only been shown on specially designed
hardware~\cite{cnvlutin,relusparse}. Structured spatial sparsity, on the other hand, can be made
efficient by using a sparse convolution operator~\cite{perforatedcnn,sbnet,submanifold}, which in turn allows the
network to shift its focus on more difficult parts of the
inputs~\cite{adaptivecomp,nopixelequal,sbnet,pag}. In self-driving, \cite{prioritize} proposed a ranking
function to prioritize computations that would have the most impact on motion planning. Weight
pruning~\cite{sparsecnn,netslim} is another popular way to achieve sparsity in the parameter space,
which is an orthogonal direction to our  method.

\textbf{Attention and loss weighting in multi-task learning:}
Our end-to-end self-driving network is an instance of multi-task learning as all three
tasks---perception, prediction and motion planning---are simultaneously solved by individual output
branches with shared features. It is common to use a summation of all the loss functions, but
sometimes there are conflicting objectives among the tasks. Prior literature in multi-task learning
has studied dynamic weighting towards different loss components, by using training signals such as
uncertainty~\cite{mtluncertain}, gradient norm~\cite{gradnorm}, difficulty
level~\cite{dynamicprioritize}, or entirely data-driven objectives~\cite{adaptiveweight,l2rw}. In
\cite{adaptiveweight,l2rw}, task and example weights are learned by optimizing the performance of the
main task. The attention mechanism has also been used in multi-task learning: in \cite{e2emtl}, a
network applies task-specific attention masks on shared features to encourage the outputs to be more
selective. Similar to dynamic loss weighting models~\cite{adaptiveweight}, we exploit the learned
attention towards weighting instance detection losses. Instead of using multiple attentions, as was
done in~\cite{e2emtl}, we use one single attention mask to optimize our main task: driving.

\textbf{Safety-driven learnable motion planning:}
One of the primary motivations of introducing attention into an end-to-end motion planning network is
to improve safety. Traditionally, safety for self-driving models was done in terms of
formal model checking and validation~\cite{formalsafety,combinatorialsafe,setsafety,failsafe,pnpsim}. More
recently, with the widely available driving data, imitation learning has been introduced in
self-driving to learn from cautious human driving~\cite{nmp,baidu,jointplt,pthree,dsdnet}. Safety has also been
considered in terms of explicitly learning a risk-sensitive measure from human
demonstration~\cite{riskirl,riskgail}. In our work, although safety is not explicitly encoded in our
loss function, we have experimentally verified that the sparse attention models are significantly
better at avoiding collisions.
% !TEX root = ../main.tex
\begin{figure}[t]
  \centering
  \includegraphics[width=\linewidth,trim={1cm 12.5cm 5cm 0},clip]{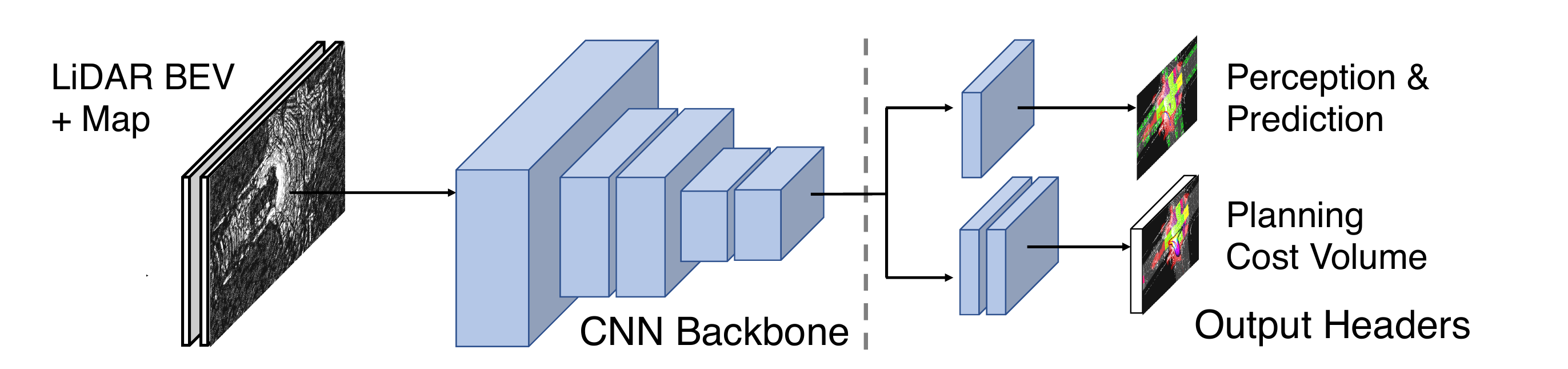}
  \caption{\small The full neural motion planner (NMP) \cite{nmp} with backbone network and header networks.}
  \label{fig:nmp}
  \vspace{-0.15in}
\end{figure}
\section{Perceive, Attend and Drive}
In this section, we present our framework for using learned, motion-planning aware attention.
We first describe  the end-to-end neural motion planner that serves as the
starting point of our work, and then introduce our proposed attention module and
attention-driven loss function, which enable us to focus the computation in areas that matter for the end task of driving.

\begin{figure}[t]
  \centering
  \includegraphics[width=\linewidth,trim={1.2cm 9cm 0 0},clip]{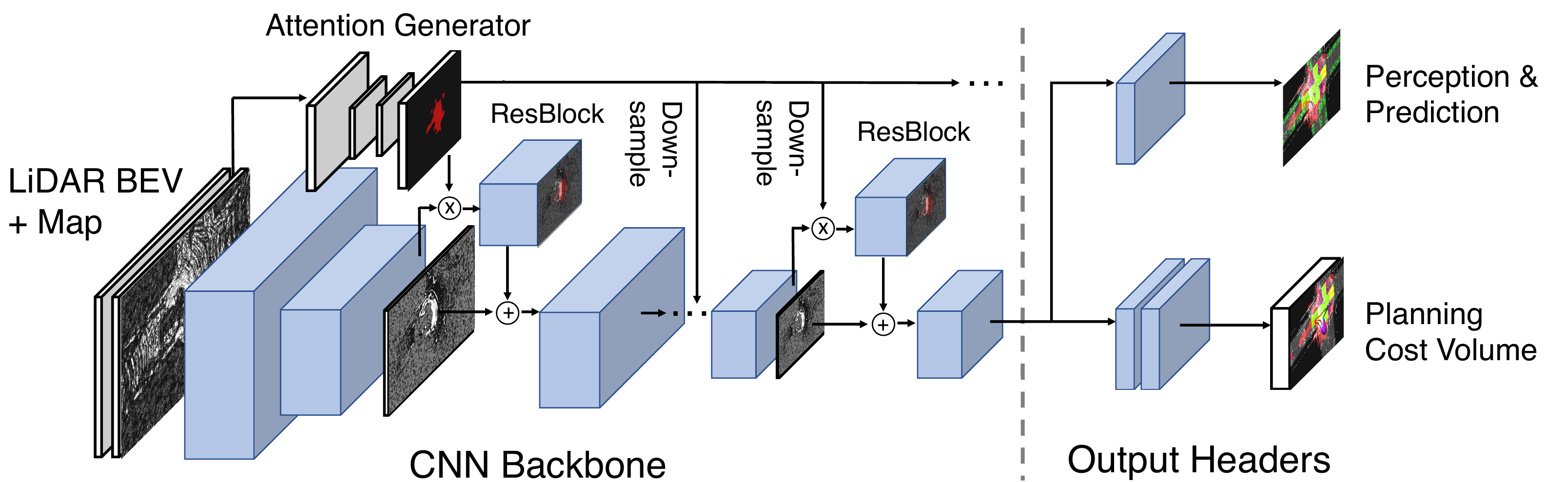}
  \caption{\small Our sparse attention neural motion planner (SA-NMP), which takes in LiDAR and HDMaps
  data, and outputs perception, prediction, and planning. Attention is generated from a U-Net and
  applied on the input branches of residual blocks. Our attention is learned towards the planning task, which is a direct model output.}
  \label{fig:mainfig}
  \vspace{-0.15in}
\end{figure}

\subsection{A review on Neural Motion Planner (NMP)}
Our proposed model extends upon the neural motion planner (NMP), which jointly solves the
perception, prediction and planning problems for self-driving. In this section we briefly review NMP depicted in Figure~\ref{fig:nmp}, and refer the reader to ~\cite{nmp} for more details.

\textbf{Input and backbone:} NMP voxelizes LiDAR point clouds to a birds-eye-view (BEV) feature map and fuses them with $M$ channels of rasterized HDMap features to produce an input representation of size $(ZT'+M)\times H\times W$, where $Z,H,W$ are the height and spatial dimensions and $T'=10$ is the number of input LiDAR sweeps. The backbone consists of 5 blocks, with the first 4 producing multi-scale features that are concatenated and fed to the final block. Overall the backbone downsamples the spatial dimension by 4.

\textbf{Multi-task headers:}
Given the features computed by the backbone, $\Inp \in \mathbb{R}^{128\times \frac{H}{4} \times \frac{W}{4}}$, NMP uses two separate headers for perception \& prediction, and motion planning.
 The perception \& prediction header consists of separate branches for classification and regression. The classification branch outputs a score for each anchor box at each spatial location over the feature map $X$, while the regression branch outputs regression targets for each anchor box, including targets for localization offset, size, and heading angle.
The planning header consists of convolution and deconvolution layers to produce a cost volume $ \cC \in \mathbb{R}^{T \times H \times W}$ representing the cost for the self-driving vehicle to be at each location and time, with $T$ the fixed future planning horizon.

\textbf{Planning inference:}
At inference time, NMP samples $N$ trajectories  that are physically realizable, and chooses the lowest cost trajectory for the ego-car:
\vskip -0.15in
\begin{align}
\tau^\star(\Inp) = \argmin_{\tau_{1 \dots N}} \ \ c(\tau_i, \Inp),
\end{align}
% \vskip -0.1in
where the cost of a
trajectory $\tau = (x_t, y_t)_{t=1}^{T}$ is the sum of all waypoints in the cost volume:
\vskip -0.15in
\begin{align}
c(\tau, \Inp) = \sum_{t=1}^T \cC_{t, x_t, y_t}(\Inp).
\end{align}
% \vskip -0.05in
We sample trajectories using a mixture of Clothoid~\cite{clothoid}, circle, and straight curves. We refer the readers to \cite{nmp} for more details on the sampling procedure.

\subsection{Sparse Attention Neural Motion Planner (SA-NMP)}
% \vspace{-0.1in}
In this section, we propose our sparse spatial attention module for self-driving, shown in Fig.~\ref{fig:mainfig}, which learns to save computation while performing well on the end task of driving safely to the goal.

\textbf{Input and backbone:} We exploit the same input representation as NMP and use the same perception, prediction, and planning headers.
The NMP backbone is replaced with the state-of-the-art backbone network of  PnPNet~\cite{pnpnet}, which uses cross-scale blocks throughout to fuse BEV sensory input. Each cross-scale block consists of three parallel branches at different resolutions that downsample the feature map, perform bulk computations, and then upsample back to the backbone resolution, before finally fusing cross-scale features across all branches. There is an additional residual connection across each cross-scale block. The final output feature from the backbone consists of 128 channels at 4$\times$ downsampled resolution, which is forwarded to the planning and detection headers.
In addition to the improved performance, PnPNet can be easily scaled for different computational budgets by varying the depth and width of the cross-scale blocks.

Existing attention-driven approaches~\cite{sbnet} tackle only the perception task and use either a road mask obtained from map information or a vehicle mask produced by a different perception module. As a consequence, they waste computation on areas that will not affect the self-driving car. We instead propose a novel approach that is end-to-end trainable and performs computation selectively for planning a safe maneuver.
As shown in Fig.~\ref{fig:mainfig}, the learned attention mask then gates the backbone network, limiting computation to areas where attention is active. By
using binary attention, we can leverage sparse convolution to improve the computational efficiency.

\textbf{Generating binary attention:}
Computational efficiency has been shown to be one of the most prominent advantages of using the
attention mechanism. For soft attention masks the computation is still dense across the entire
activation map, and therefore no computation savings can be achieved.  SBNet~\cite{sbnet} showed that a sparse convolution operator can achieve significant speed-ups with a given
discrete binary attention mask. Here, we would also like to exploit the computational benefit of
sparse convolution by using discrete attention outputs.

We utilize a network to
predict a scalar score for each spatial location, and binarize the score to represent our sparse attention map. For efficiency and simplicity, we use a small U-Net~\cite{unet} with skip
connections and two downsample/upsample stages.
We would like to apply the generated attention back to the BEV features in the
model backbone so as to sparsify the spatial information, allowing computation to be focused on the
important regions only. We choose to do so in a residual manner~\cite{resattn,sbnet} to avoid
deteriorating the features throughout the backbone. Let $x + F(x)$ denote the normal residual block.
Our attention mask is multiplied with the input to the residual block as follows:
\begin{align}
\mathrm{ResAttend}(x) = x + F(x \odot A),
\end{align}
where $\odot$ denotes elementwise multiplication. See Fig. \ref{fig:mainfig} for an illustration of our architecture.

\textbf{Learning binary attention with Gumbel softmax:}
In order to learn the attention generator and backpropagate through the binary attention map, we make use of the Gumbel
softmax technique \cite{gumbel,concrete} since the step function is not differentiable, and using
the standard sigmoid function suffers from a more severe bias-variance trade-off~\cite{gumbel}. Let
$i,j$ denote spatial coordinates, and $z_{i,j}$ the scalar output from the attention U-Net. We first
add the Gumbel noise on the logits as follows:
\vskip -0.15in
\begin{align}
  \pi_{i,j} &= \sigmoid(z_{i,j}) \\
  \alpha_{i,j}^{(0)} &= \log \pi_{i,j} + g_{i,j}^{(0)}\\
  \alpha_{i,j}^{(1)} &= \log (1 - \pi_{i,j}) + g_{i,j}^{(1)},
\end{align}
where $g_{i,j} = -\log(-\log u)$, and $u$ is sampled from $\textrm{Uniform}[0,1]$. At inference time,
hard attention $A_{i,j}$ can be obtained by comparing the logits,
\begin{align}
\attn_{i,j} &= \begin{cases}
1 & \text{if} \ \ \ \alpha_{i,j}^{(0)} \ge \alpha_{i,j}^{(1)} \\
0 & \text{otherwise}.
\end{cases}
\end{align}
During training, however, we would like to approximate the gradient by using the straight-through
estimator~\cite{gumbel,ste}. Hence, in the backward pass, the step function is replaced with a
softmax function with a temperature constant $\temp$ (where $\sattn$ is the underlying soft attention):
\begin{align}
\sattn_{i,j} &= \frac{\exp \left( \alpha_{i,j}^{(0)}/\temp \right)}
  {\exp\left(\alpha_{i,j}^{(0)} / \temp \right)
   + \exp\left(\alpha_{i,j}^{(1)} / \temp \right)}.
\end{align}

\subsection{Multi-Task Learning}
\label{sec:loss}
We train our sparse  neural motion planner (including the attention) end-to-end with a multi-task learning
objective that combines planning ($\cL_{\text{plan}}$) with perception \& motion forecasting ($\cLcla$, $\cLreg$):
\begin{align}
\cL = \lpln \cLpln + \lcla \cLcla + \lreg \cLreg + \lalp \cLatn + \lambda \lVert w \rVert^2_2,
\label{eq:joint}
\end{align}
where $\cLatn$ is an $\ell_1$ loss, defined in Equation~\ref{eq:regularize}, that controls the sparsity of the attention mask, and $\lVert w
\rVert^2_2$ is the standard weight decay term. Following \cite{nmp, pnpnet}, we fix $\lpln=0.001, \lcla=1.0,
\lreg=0.5$.

\textbf{Motion planning loss:} The motion planning loss utilitizes the max-margin objective, where
the ground-truth driving trajectory (performed by a human) should be of lower cost than other
trajectories sampled by the model. Let $(x_t, y_t)$ be the groundtruth trajectory and let $c_t$ be
the cost volume output by the model for timestamp $t$. We randomly sample $N$ trajectories serving as negative samples: $\{x_t^{(i)}, y_t^{(i)}\}_{i=1}^N$, and  penalize the maximum margin violation between groundtruth and negative samples:
\begin{align}
\cLpln &= \max_{i=1 \dots N} \sum_{t=1}^{T} \max\{ 0, c_t - c^{(i)}_t + \Delta_t^{(i)} \},
\end{align}
where $\Delta^{(i)}_t$ is the task loss capturing spatial differences, and traffic violations denoted by $v$. $v^{(i)}_t$ is non-zero when negative trajectory $i$ violates a traffic rule at time $t$:
\begin{align}
\Delta^{(i)}_t &= \left \lVert (x_t, y_t) - (x^{(i)}_t, y^{(i)}_t) \right \rVert_2 + v^{(i)}_t.
\end{align}

\textbf{Perception \& prediction (PnP) loss:} This loss follows the
standard classification and regression objectives for object detection. The classification part uses binary cross-entropy:
\begin{align}
\hspace{-0.07in}
\cLclaij = \sum_k -\hat{y}_{i,j,k} \log(y_{i,j,k}) - (1- \hat{y}_{i,j,k}) \log (1 - y_{i,j,k}),
\end{align}
where $y$ is the predicted classification score between 0 and 1, and $\hat{y}$ is the binary
ground truth. For each detected instance, the model outputs a bounding box, and a pair of coordinates
and angles for each future step. We reparameterize the shift of a bounding box $(x, y, w, h, \theta)$ from
an anchor bounding box $(x_a, y_a, w_a, h_a, \theta_a)$ in a 6-dimensional vector $\delta$:
\begin{align}
\delta_t = 
[
\frac{x_a - x}{w_a},
\frac{y_a - y}{h_a},
\log \frac{w}{w_a},
\log \frac{h}{h_a},
\sin (\theta_a - \theta),
\cos (\theta_a - \theta)
].
\end{align}
A regression loss is then applied for the trajectory of the instance up to time $T$. For each
spatial coordinate $(i,j)$, we sum up the losses of all bounding boxes $b$ where the ground truth belongs to this location:
\begin{align}
\cLregij = \sum_{b \in (i, j)} \sum_{t=0}^{T} \mathrm{SmoothL1}(\hat{\delta}_{b,t}, \delta_{b,t}),
\end{align}
with $\hat{\delta}$  the predicted shifts and $\delta$  the ground truth shifts.

\textbf{Auxiliary loss masking:} Our overall objective is to achieve good performance
in motion planning, so PnP is an auxiliary task. Since the majority of our
computation happens within the attended area, intuitively the model should not be penalized as
severely for mis-detecting objects not in the attended area. We, therefore, propose to use
our spatial attention mask $A$ to re-weight the PnP losses as follows:
\begin{align}
\cLcla &= \gamma_1 \sum_{i,j} \attn_{i,j} \cLclaij + \gamma_0 \sum_{i,j} \cLclaij,\\
\cLreg &= \gamma_1 \sum_{i,j} \attn_{i,j} \cLregij + \gamma_0 \sum_{i,j} \cLregij,
\label{eq:reweight}
\end{align}
where $\gamma_1$ weights attended instances, and $\gamma_0$ weights all instances. We fix $\gamma_0 = 0.1$ and $\gamma_1 = 0.9$.

\textbf{Attention sparsity loss:}
To encourage focused attention and high sparsity, we use an $\ell_1$ regularizer on the attention mask as follows. We control sparsity with $\lambda_A$ in Equation~\ref{eq:joint}.
\vskip -0.05in
\begin{align}
\cLatn = \sum_{i,j} \attn_{i,j}.
\label{eq:regularize}
\end{align}
\vskip -0.05in
% !TEX root = ../main.tex
\begin{figure*}[t]
\centering
\includegraphics[width=0.95\textwidth]{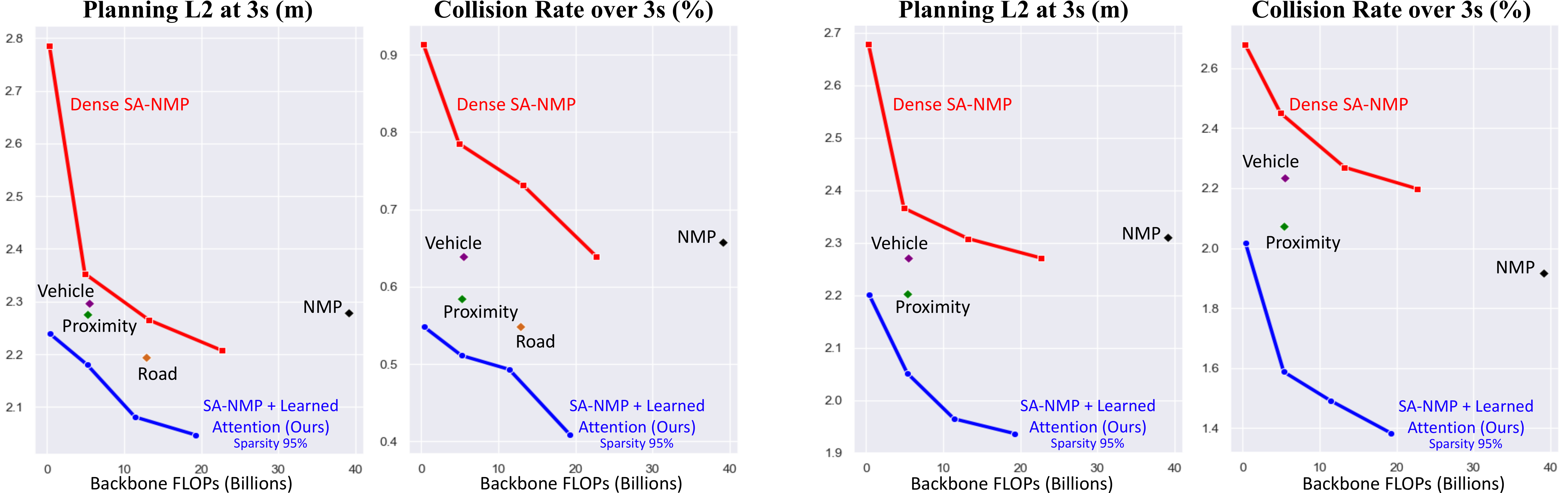}
\caption{\small Planning performance of our learned sparse attention model compared to other baselines at varying computation budgets (lower is better on both metrics). \textbf{Left}: \ourdata{}; \textbf{Right}: nuScenes. Note that all models except for \textit{NMP} use the same SA-NMP backbone, which can be scaled by changing the depth and width, allowing the computation of \textit{Dense SA-NMP} and \textit{SA-NMP+Learned Attention} to be varied.
}
\label{fig:mainresult}
\end{figure*}

\section{Experimental Evaluation}
We evaluated on a 
real-world driving dataset (\ourdata{}), training on over 1 million frames from 5,000 scenarios and validating on 5,000 frames from 500 scenarios, using both LiDAR and HD-maps.
We also evaluated on  nuScenes v1.0~\cite{nuscenes}, a large-scale public dataset, with a
training set of over 200,000 frames and a test set of 5,000 frames. Due to the inaccurate localization they provide, we omitted HDMaps and only used LiDAR~\cite{pnpnet}.

\subsection{Implementation Details and Metrics}
\label{sec:impl}

\textbf{Training:}
To jointly train SA-NMP with attention, we use pretrained
weights for the backbone and headers from training a SA-NMP
without attention (dense) for two epochs. 
We train all our models with batch size 5 across 16 GPUs in parallel using the Adam \cite{adam} optimizer. We use an initial learning rate of $1 \times 10^{-4}$, and decay of 0.1 at 1.0 and 1.6 epoch(s), for a total of 2.0 epochs.

\textbf{Evaluation:}
To evaluate driving and safety performance, we focus on the following planning metrics which are accumulated over all 6 future timesteps (3s): \textit{Planning L2} is the L2 distance between waypoints of the predicted future ego trajectory and those of the ground-truth trajectory (characterized by human driving). \textit{Collision rate} is the frequency of collisions between the planned ego trajectory and the ground truth trajectories of other actors in the scene. \textit{Lane violation rate} measures the number of lane boundary violations by the planned ego trajectory. 
We do not evaluate this on nuScenes due to the inaccurate localization they provide,

\textbf{Baselines:}
We compare our learned attention to baselines that are end-to-end trained using static attention masks obtained from priors. 
\textit{Road Mask} covers the entire road as provided from the map data. \textit{Vehicle Mask} strictly covers all detections in the input space, obtained from a PSPNet \cite{pspnet} trained for segmentation. \textit{Proximity Mask} is a circular radius around the ego vehicle. \textit{Dense} is
not using sparse attention.

% !TEX root = ../main.tex

% !TEX root = ../main.tex
\setlength{\tabcolsep}{6pt}
\begin{table*}[t]
\centering
\caption{\small Performance and efficiency of our learned attention model vs. dense and simple attention baselines.
}
\resizebox{0.7\linewidth}{!}{
\begin{tabular}{lccccc}
\hline

\multirow{2}{*}{\bf{\ourdata{}}} & \bf{Backbone} & \bf{Sparsity} & \bf{Planning L2} & \bf{Collision Rate} & \bf{Lane Violation}\\
& \bf{FLOPS} & & \bf{at 3s (m)} & \bf{over 3s (\%)} & \bf{over 3s (\%)} \\
\hline
NMP \cite{nmp}            & 39.18B      &  0.0\%      & 2.279      & 0.657      & 2.780      \\
Dense SA-NMP              & 22.73B      &  0.0\%      & 2.207      & 0.639      & 1.350      \\
\hline
SA-NMP+Vehicle Mask        & 5.43B       & 93.6\%      & 2.297      & 0.639      & 1.387      \\
SA-NMP+Proximity Mask      & 5.31B       & 94.0\%      & 2.276      & 0.584      & 1.387      \\
SA-NMP+Road Mask           & 12.85B      & 68.9\%      & 2.194      & 0.548      & \bf{1.296} \\
SA-NMP+Learned Attn (Ours) & \bf{5.22B}  & \bf{95.0\%} & \bf{2.102} & \bf{0.511} & 1.338      \\
\hline
\end{tabular}}
\\
\resizebox{0.63\linewidth}{!}{
\begin{tabular}{lccccc}
& & & & & \\
\hline
\multirow{2}{*}{\bf{nuScenes}} & \bf{Backbone} & \bf{Sparsity} & \bf{Planning L2} & \bf{Collision Rate} \\
 & \bf{FLOPS} & & \bf{at 3s (m)} & \bf{over 3s (\%)} \\
\hline
NMP \cite{nmp}             & 39.18B      &  0.0\%      & 2.310      & 1.918      \\
Dense SA-NMP               & 22.73B      &  0.0\%      & 2.271      & 2.198      \\
\hline
SA-NMP+Vehicle Mask        & 5.44B       & 94.0\%      & 2.263      & 2.234      \\
SA-NMP+Proximity Mask      & \bf{5.31B} & 94.0\% & 2.103      & 2.073      \\
SA-NMP+Learned Attn (Ours) & 5.34B      & 94.0\% & \bf{2.052} & \bf{1.588} \\
\hline
\end{tabular}}
\label{tab:main}
\vspace{-0.2cm}
\end{table*}

\subsection{Results}

\textbf{Quantitative results:} With a sparse attention mask learned towards motion planning, we can leverage the sparsity in the network backbone to greatly reduce computational costs, while not only maintaining but improving model performance. In our experimental results, we use theoretical FLOPs to show the efficiency of our network, but this also translates to realtime gains as SBNet \cite{sbnet} has been shown to leverage sparsity to achieve real speed-ups.
The increase in efficiency from leveraging sparsity is shown in Table~\ref{tab:main}, where our learned attention model uses $\sim80\%$ fewer FLOPs than \textit{Dense SA-NMP} thanks to its 95\% sparse attention mask. Also, even with an identical SA-NMP backbone as the baselines (except \textit{NMP}), our model with
learned attention performs better in all motion planning metrics, which indicates that focused backbone computation is greatly advantageous to the overall goal of safe planning. \textit{NMP+Road} performs slightly better in \textit{Lane Violation} due to the road mask attention focusing on all road/lane markings. However, this baseline uses more than double the FLOPs since its attention mask looks at the entire road surface at only 68.9\% sparsity. From Fig.~\ref{fig:mainresult}, our learned attention model clearly outperforms other baselines in collision rate and planning L2, across all computational budgets, by varying the depth and width of the backbone network.

\begin{figure*}[t]
\centering
  \includegraphics[width=\linewidth]{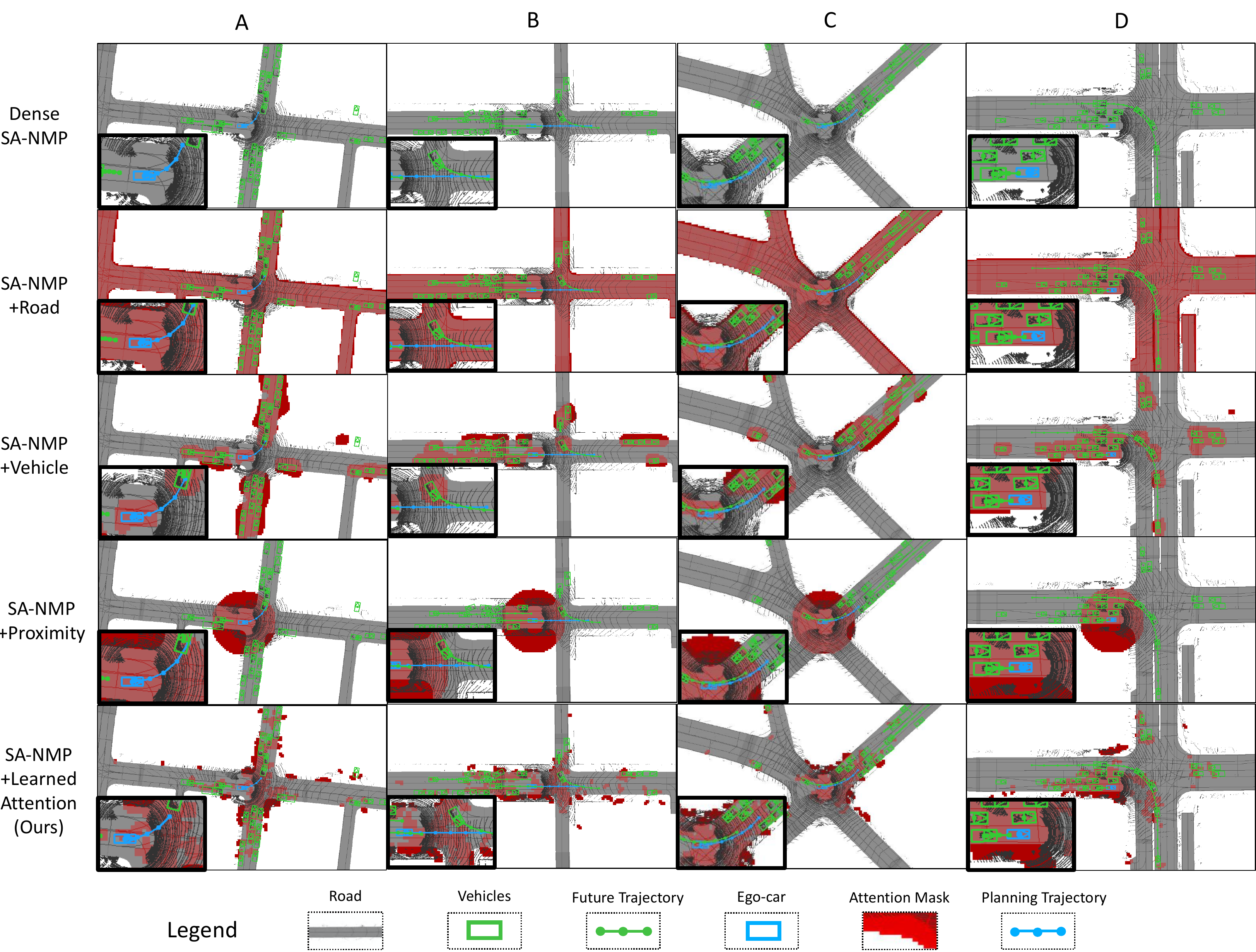}
  \vspace{-0.2in}
  \caption{\small Visualization of the attention masks and planned trajectory comparing dense, road mask, vehicle mask, proximity mask, and our learned attention. %Each is applied to SA-NMP. 
  \textbf{Col A:} baselines turn too fast and collide with vehicle ahead. \textbf{B:} baselines collide with the future position of a left-turning vehicle. \textbf{C:} a tight left-turn where all models collide with or nearly miss parked vehicles. \textbf{D:} a
  rear-end collision for all models.}
  \label{fig:joint_vis}
  \vspace{-0.1in}
\end{figure*}

\textbf{Qualitative results:}
In Figure~\ref{fig:joint_vis}, we show examples of our learned attention compared to baselines.
As expected, our model focuses on the road and vehicles directly ahead; however,
it also diverts some amount of attention to distant vehicles and road markings. This ability
to dynamically distribute attention is likely why our model outperforms the baselines,
which attend either indiscriminately (Dense and Road) or too selectively (Vehicle and Proximity). From the visualizations, we can better understand our model's improved collision avoidance. Since the attention is dynamic, our model is more effective at anticipating other vehicles resulting in more cautious planning. This is illustrated by Columns A, B in Fig.~\ref{fig:joint_vis}, where our planned trajectory avoids future collisions with others. 
The failure cases mostly arise from rear-end collisions, one of which is Column D where all models are hit by the trailing vehicle. Note that this arises as we evaluate in open-loop. Our model focuses on surrounding vehicles and not enough on the open road to its right, which would give the option of making a right turn.

\textbf{Sparsity of learned attention:} Table~\ref{tab:ablation_sparse} shows the result of varying $\lambda_A$ from Eq.~\ref{eq:joint}, which weights the $\ell_1$ regularization term from Eq.~\ref{eq:regularize}, with other settings held constant. We found that overall motion planning performance improves with increased sparsity, or essentially more focused computation, and peaks at 95\% sparsity. 

% !TEX root = ../main.tex
\setlength{\tabcolsep}{6pt}
\begin{table}
\centering
    \caption{\small Varying learned attention sparsity with $\lambda_A$.}
    \vspace{-0.1in}
    \resizebox{0.48\textwidth}{!}{
    \begin{tabular}{lcccc}
      \hline
      \multirow{2}{*}{\bf{Sparsity}}  & \multirow{2}{*}{$\lambda_A$}  & \bf{Planning L2} & \bf{Collision Rate} & \bf{Lane Violation}\\
      & & \bf{at 3s (m)} & \bf{over 3s (\%)} & \bf{over 3s (\%)} \\
    \hline
      Dense 0\% & -           & 2.207      & 0.639      & 1.350      \\
      \hline
      Ours 20\% & $1.0 \times 10^{-8}$ & 2.179      & 0.547      & 1.352      \\
      Ours 50\% & $1.0 \times 10^{-7}$ & 2.138      & 0.620      & 1.361      \\
      Ours 75\% & $5.0 \times 10^{-7}$ & 2.132      & 0.584      & 1.387      \\
      Ours 95\% & $1.0 \times 10^{-6}$ & \bf{2.102} & \bf{0.511} & \bf{1.338} \\
      Ours 99\% & $5.0 \times 10^{-6}$ & 2.211      & 0.566      & 1.367      \\
    \hline
    \end{tabular}
    }
    \label{tab:ablation_sparse}
  \vspace{-0.1in}
\end{table}
\begin{table}

  \centering
    \caption{\small Influence of loss reweighting ratio $\gamma_1$.}
    \vspace{-0.1in}
    \resizebox{0.4\textwidth}{!}{
      \begin{tabular}{lccc}
      \hline
        \multirow{2}{*}{\bf{Ratio}} 
        & \bf{Planning L2} & \bf{Collision Rate} & \bf{Lane Violation}\\
        & \bf{at 3s (m)} & \bf{over 3s (\%)} & \bf{over 3s (\%)} \\
        \hline
        $\gamma_1=1.00$   & 2.210      & 0.548      & 1.378      \\
        $\gamma_1=0.90$   & \bf{2.102} & \bf{0.511} & \bf{1.338} \\
        $\gamma_1=0.75$ & 2.230      & 0.621      & 1.393      \\
        $\gamma_1=0.50$   & 2.194      & 0.637      & 1.405      \\
        $\gamma_1=0.00$   & 2.188 & 0.633 & 1.397 \\
        \hline
      \end{tabular}}
    \label{tab:ablation_reweight}
  \vspace{-0.25in}
\end{table}

\textbf{Perception and prediction (PnP) loss reweighting:} Table~\ref{tab:ablation_reweight} shows results with varying $\gamma_1$ and $\gamma_0 = 1 - \gamma_1$ from Eq.~\ref{eq:reweight} which control the weighting of the PnP loss computed on actors inside vs. outside the attention mask. All other variables are fixed, including sparsity at 95\%. 
As $\gamma_1$ increases, the learned attention is less restricted by detection performance on all actors, and is able to focus on only the most important actors and parts of the road, distributing attention towards improving motion planning performance. Note that $\gamma_1=1.0$ is an extreme case where PnP loss is computed only on actors within attention mask: the model learns to cheat by generating attention that avoids all actors resulting in no PnP learning signal, hence the poor performance. For our main experiments, we fix $\gamma_1=0.9$.

\textbf{Detection performance:}
Since the overall goal is improved motion-planning with lighter computation, focusing on accurately detecting all actors indiscriminately would contradict the purpose of our learned sparse attention. 
We should not care as much about far away or irrelevant actors that have no effect on safe planning, and should instead focus our computation on important input regions.
Table~\ref{tab:perception} compares detection performance between our learned attention and the baseline dense model evaluated on different subsets of actors in the scene. 
\textit{Full} includes all actors in the input, while 
\textit{Attended Region} is the subset of actors that lie within the attention mask. For evaluating the dense model, we use the attention mask generated by our learned model to get the \textit{Attended Region}, ensuring that both models are evaluated on the same actor subsets in both settings. 
The results show that our 95\% sparse, learned attention model is better than the dense model at detecting actors within the attention mask, meaning that its performance is better focused on actors that it believes are important. 
This may explain the overall improved planning performance of our attention-driven models as demonstrated in the main quantitative and qualitative results.

% !TEX root = ../main.tex
\setlength{\tabcolsep}{6pt}
\begin{table}[t]
\caption{\small Detection performance on different input regions.}
\centering
\begin{small}
\begin{center}
\resizebox{0.98\columnwidth}{!}{
\begin{tabular}{lccc|ccc}
% \bf{Model} 
\hline
& \multicolumn{3}{c|}{\bf{mAP on Full}} &\multicolumn{3}{c}{\bf{mAP on Attended Region}}\\
             & \bf{IoU@0.3}   & \bf{IoU@0.5}   & \bf{IoU@0.7}   & \bf{IoU@0.3}   & \bf{IoU@0.5}   & \bf{IoU@0.7}     \\
\hline
Dense SA-NMP         & 
\bf{97.8} & \bf{94.7} & \bf{80.3} & 94.1      & 93.3      & 87.9        \\
Ours 95\% Sparse  & 96.3      & 92.1      & 74.9      & \bf{94.2} & \bf{93.8} & \bf{88.5}   \\
\hline
\end{tabular}
}
\end{center}
\end{small}
\vspace{-0.3in}
\label{tab:perception}
\end{table}

% !TEX root = ../main.tex
\vspace{-0.1in}
\section{Conclusion}
In this work, we propose an end-to-end learned, sparse visual attention
mechanism for self-driving, where the sparse attention mask gates the feature backbone
computation. As opposed to existing methods that focus on using attention for perception only,
our attention masks are directly optimized for motion planning, which enables our network to output
better planned trajectories while achieving more efficiency with higher sparsity. In future work,
the attention module can be extended to have recurrent feedbacks from the output layers
to better leverage temporal information.

\bibliographystyle{IEEEtran}
\bibliography{refs}

\end{document}